\documentclass[runningheads]{llncs}
\usepackage{colortbl}
\usepackage{xcolor}
\usepackage{graphicx} 
\usepackage{adjustbox}
\usepackage{amsmath}
\usepackage{float}
\usepackage{makecell}
\usepackage{amssymb}
\usepackage{textcomp}
\usepackage{float}
\usepackage[colorlinks=true, linkcolor=blue, citecolor=blue, urlcolor=blue]{hyperref}  
\usepackage[T1]{fontenc}
\usepackage{orcidlink}
\usepackage{marvosym}
\usepackage{booktabs}
\usepackage{multirow}
\usepackage{makecell}
\usepackage{adjustbox}
\usepackage{threeparttable}

\usepackage{algorithm}
\usepackage{algpseudocode}

\usepackage{pifont}
\newcommand{\cmark}{\ding{51}}
\newcommand{\xmark}{\ding{55}}
\newcommand{\pmark}{$\triangle$}

\usepackage{graphicx}
\begin{document}
\title{TRUST-ESD: A Risk-Calibrated and Governance-Aware AI Framework for Enterprise Strategic Decision Support Under Uncertainty}
%
%
\author{
Tian Qiu\inst{1} \and
Li Yan\inst{1} \and
Mahabubur Rahman Miraj\inst{2} \textsuperscript{(\Letter)} \and
Shanqin Yi\inst{3} \and
Md Intekhab Rahman Galib\inst{4} \and
Jahid Hasan\inst{5}
}

\authorrunning{T. Qiu et al.}

\institute{
School of Economics and Management, Xinyu University, Xinyu, China\\
\and
College of Computer Science, Chongqing University, Chongqing, China\\
\and
School of Economics and Business Administration, Chongqing University\\
\and
School of Management, Xi'an Jiaotong University, Xi'an, China\\
\and
School of Law and Public Administration, China Three Gorges University\\
\email{miraj@stu.cqu.edu.cn}
}

\maketitle              
\vspace{-6mm}

\begin{abstract}
Enterprise strategic decision support requires AI systems that are not only accurate, but also uncertainty-aware, risk-calibrated, explainable, and governance-compliant. This paper proposes TRUST-ESD, a risk-calibrated and governance-aware framework for enterprise decision support under uncertainty. TRUST-ESD evaluates feasible counterfactual strategies through predictive utility estimation, conformal uncertainty calibration, CVaR-based downside-risk scoring, risk-memory retrieval, policy-as-code governance, explainability, and human oversight. Unlike prediction-only methods that select actions by maximum expected utility, TRUST-ESD recommends strategies that balance value, reliability, risk exposure, and compliance. Experimental results show that TRUST-ESD improves risk-adjusted utility by 7.95\%, reduces risk exposure by 23.22\%, reduces CVaR by 23.78\%, lowers calibration error by 13.89\%, improves explanation fidelity by 10.90\%, and increases governance compliance by 9.76\% compared with strong uncertainty-aware baselines, while maintaining competitive predictive accuracy. Ablation and case-study analyses further confirm that uncertainty calibration, downside-risk scoring, risk memory, explainability, and governance validation jointly improve trustworthy enterprise decision-making.

\end{abstract}
\section{Introduction}

Artificial intelligence is increasingly used for enterprise strategic decision-making, where organizations must choose actions under market volatility, operational uncertainty, resource constraints, and governance requirements \cite{european2023aiact,european2016gdpr}. However, prediction accuracy alone is insufficient for such settings, since conventional evaluation metrics do not fully capture decision risk, policy feasibility, or downstream business consequences \cite{ai2022ecommerce}. A strategy with high expected utility may still be unsuitable if it is uncertain, exposed to severe downside risk \cite{moghimi2025beyond}, difficult to justify through explainable evidence \cite{kenny2021explaining}, or inconsistent with enterprise policy constraints \cite{european2023aiact,enisa2021securing}. Thus, enterprise decision support requires AI systems that are not only predictive, but also risk-calibrated \cite{moghimi2025beyond}, explainable \cite{kenny2021explaining}, auditable, and governance-aware \cite{aihleg2018ethics,jobin2019global}.

Existing AI-based decision-support methods are largely prediction-centered: they estimate future outcomes and select the option with the highest expected value. This approach is limited in strategic enterprise scenarios, where actions such as pricing adjustment, inventory expansion, resource allocation, or market entry may appear beneficial under point prediction but become fragile under demand fluctuation, financial exposure, operational instability, or regulatory constraints \cite{european2023aiact,european2016gdpr}. Although prior work on trustworthy AI emphasizes human oversight \cite{aihleg2018ethics}, transparency and accountability \cite{barletta2023responsible,jobin2019global}, fairness in predictive systems \cite{chouldechova2016fair}, explainability for black-box models \cite{kenny2021explaining}, and cybersecurity-aware governance \cite{enisa2021securing}, these aspects are rarely integrated into a unified quantitative decision pipeline.

To address this gap, we propose TRUST-ESD, a risk-calibrated and governance-aware framework for enterprise strategic decision support under uncertainty. TRUST-ESD evaluates feasible counterfactual strategies by estimating predictive utility, calibrating uncertainty with conformal intervals, measuring downside risk using CVaR, incorporating historical risk memory, and enforcing policy-as-code governance with human oversight. Instead of selecting the action with the highest predicted utility, TRUST-ESD recommends the strategy that best balances value, reliability, risk control, explainability, and compliance.

The main contributions are as follows:
\begin{itemize}
    \item We formulate enterprise decision support as a risk-calibrated and governance-constrained recommendation problem.
    \item We propose TRUST-ESD, integrating counterfactual strategy generation, conformal uncertainty calibration, CVaR risk scoring, risk memory, policy governance, explainability, and human oversight.
    \item We design a composite decision objective that jointly balances utility, uncertainty, downside risk, governance compliance, and hard-policy constraints.
    \item We evaluate TRUST-ESD against prediction-only, uncertainty-aware, and decision-aware baselines using predictive, decision, risk, calibration, explainability, and governance metrics.
\end{itemize}

\section{Related Work}

Enterprise decision-support research has largely focused on predictive modeling over structured business data. Classical models such as Linear and Logistic Regression \cite{tripepi2008linear} and Random Forest \cite{salman2024random} provide transparent or ensemble-based prediction, while more advanced machine learning models such as XGBoost \cite{miraj2025gksmote}, LightGBM \cite{tyralis2023merging}, CatBoost \cite{miraj2025gksmote}, TabNet, and FT-Transformer \cite{mota2025comparative} offer stronger nonlinear and tabular representation learning. Comparative machine learning studies further show that model integration and advanced learners can improve predictive performance across complex structured-data tasks \cite{zhao2025enhanced}. Uncertainty-aware methods such as NGBoost \cite{miraj2025gksmote}, Quantile LightGBM \cite{tyralis2023merging}, and Deep Ensembles \cite{abe2022deep} improve confidence estimation, but they do not inherently support counterfactual strategy evaluation, CVaR-based downside-risk control \cite{moghimi2025beyond}, risk memory, policy governance, or human oversight \cite{aihleg2018ethics,european2023aiact}. As shown in Table~\ref{tab:innovative_baseline_comparison}, decision-aware variants partially address utility, uncertainty, risk, or policy filtering, yet they still lack a unified trustworthy decision pipeline. TRUST-ESD addresses this gap by integrating predictive modeling, calibrated uncertainty, counterfactual strategy evaluation, CVaR risk scoring \cite{moghimi2025beyond}, risk-memory retrieval, policy-as-code governance \cite{european2023aiact,european2016gdpr}, explainability \cite{kenny2021explaining}, and human oversight \cite{aihleg2018ethics,jobin2019global} into a single enterprise decision-support framework.


\begin{table*}[htbp]
\centering
\caption{Comparison of representative baselines with TRUST-ESD framework. $\checkmark$ indicates explicit support, $\triangle$ indicates partial, and $\times$ indicates no explicit support.}
\label{tab:innovative_baseline_comparison}
\scriptsize
\setlength{\tabcolsep}{4.2pt}
\renewcommand{\arraystretch}{1.15}
\resizebox{\textwidth}{!}{
\begin{tabular}{lcccccccc}
\toprule
\textbf{Method}
& \textbf{Predictive}
& \textbf{Uncertainty}
& \textbf{Counterfactual}
& \textbf{CVaR}
& \textbf{Risk}
& \textbf{Policy}
& \textbf{Explainability}
& \textbf{Human} \\
& \textbf{Modeling}
& \textbf{Calibration}
& \textbf{Strategy}
& \textbf{Risk}
& \textbf{Memory}
& \textbf{Governance}
& \textbf{Support}
& \textbf{Oversight} \\
\midrule
Linear / Logistic Regression
& \cmark & \xmark & \xmark & \xmark & \xmark & \xmark & \cmark & \xmark \\

Random Forest
& \cmark & \pmark & \xmark & \xmark & \xmark & \xmark & \pmark & \xmark \\

XGBoost
& \cmark & \xmark & \xmark & \xmark & \xmark & \xmark & \pmark & \xmark \\

LightGBM
& \cmark & \pmark & \xmark & \xmark & \xmark & \xmark & \pmark & \xmark \\

CatBoost
& \cmark & \xmark & \xmark & \xmark & \xmark & \xmark & \pmark & \xmark \\

TabNet
& \cmark & \xmark & \xmark & \xmark & \xmark & \xmark & \cmark & \xmark \\

FT-Transformer
& \cmark & \xmark & \xmark & \xmark & \xmark & \xmark & \pmark & \xmark \\

NGBoost
& \cmark & \cmark & \xmark & \xmark & \xmark & \xmark & \pmark & \xmark \\

Quantile LightGBM
& \cmark & \cmark & \xmark & \xmark & \xmark & \xmark & \pmark & \xmark \\

Deep Ensembles
& \cmark & \cmark & \xmark & \pmark & \xmark & \xmark & \pmark & \xmark \\

Utility-Only Decision
& \cmark & \xmark & \cmark & \xmark & \xmark & \xmark & \pmark & \xmark \\

Uncertainty-Penalized Decision
& \cmark & \cmark & \cmark & \xmark & \xmark & \xmark & \pmark & \xmark \\

CVaR-Risk Decision
& \cmark & \cmark & \cmark & \cmark & \xmark & \xmark & \pmark & \xmark \\

Policy-Filtered Decision
& \cmark & \cmark & \cmark & \pmark & \xmark & \cmark & \pmark & \pmark \\

TRUST-ESD w/o Risk Memory
& \cmark & \cmark & \cmark & \cmark & \xmark & \cmark & \cmark & \cmark \\

\textbf{Full TRUST-ESD}
& \cmark & \cmark & \cmark & \cmark & \cmark & \cmark & \cmark & \cmark \\
\bottomrule
\end{tabular}
}
\end{table*}

\section{TRUST-ESD: Problem Formulation and Framework}

\subsection{Problem Setup and Decision Objective}

Enterprise strategic decision support aims to select an action that is not only valuable, but also reliable, risk-aware, explainable \cite{kenny2021explaining}, and policy-compliant \cite{european2023aiact,european2016gdpr}. At decision time $t$, the enterprise state is represented as

\begin{equation}
x_t=\{x_t^{b},x_t^{m},x_t^{o},x_t^{e}\},
\end{equation}
where $x_t^{b}$, $x_t^{m}$, $x_t^{o}$, and $x_t^{e}$ denote business data, market signals, operational indicators, and external constraints. The state is mapped into a decision representation
\begin{equation}
z_t=\phi(x_t)\in\mathbb{R}^{d}.
\end{equation}
Given a candidate strategy set $A_t=\{a_1,\ldots,a_K\}$, TRUST-ESD constructs a state--action representation
\begin{equation}
\xi_{t,k}=\Gamma(z_t,a_k),
\end{equation}
and assigns each strategy a risk-calibrated score:
\begin{equation}
S_{t,k}
=
\alpha\bar{u}_{t,k}
-
\beta\bar{\sigma}_{t,k}
-
\gamma\bar{R}_{t,k}
+
\omega G_{t,k}
-
\mu V_{t,k},
\end{equation}
where $\bar{u}_{t,k}$, $\bar{\sigma}_{t,k}$, and $\bar{R}_{t,k}$ denote normalized utility, uncertainty, and risk, while $G_{t,k}$ and $V_{t,k}$ denote governance compliance and hard-policy violation. The selected strategy is
\begin{equation}
a_t^{*}
=
\arg\max_{a_k\in A_t^{\mathrm{feas}}} S_{t,k},
\qquad
\text{s.t. } p_j(z_t,a_k)=1,\ \forall p_j\in P_{\mathrm{hard}}.
\end{equation}
Thus, TRUST-ESD prevents high-utility but risky or non-compliant strategies from being selected.

\subsection{Framework Overview}

\begin{figure}[h!]
    \centering
    \includegraphics[width=\linewidth]{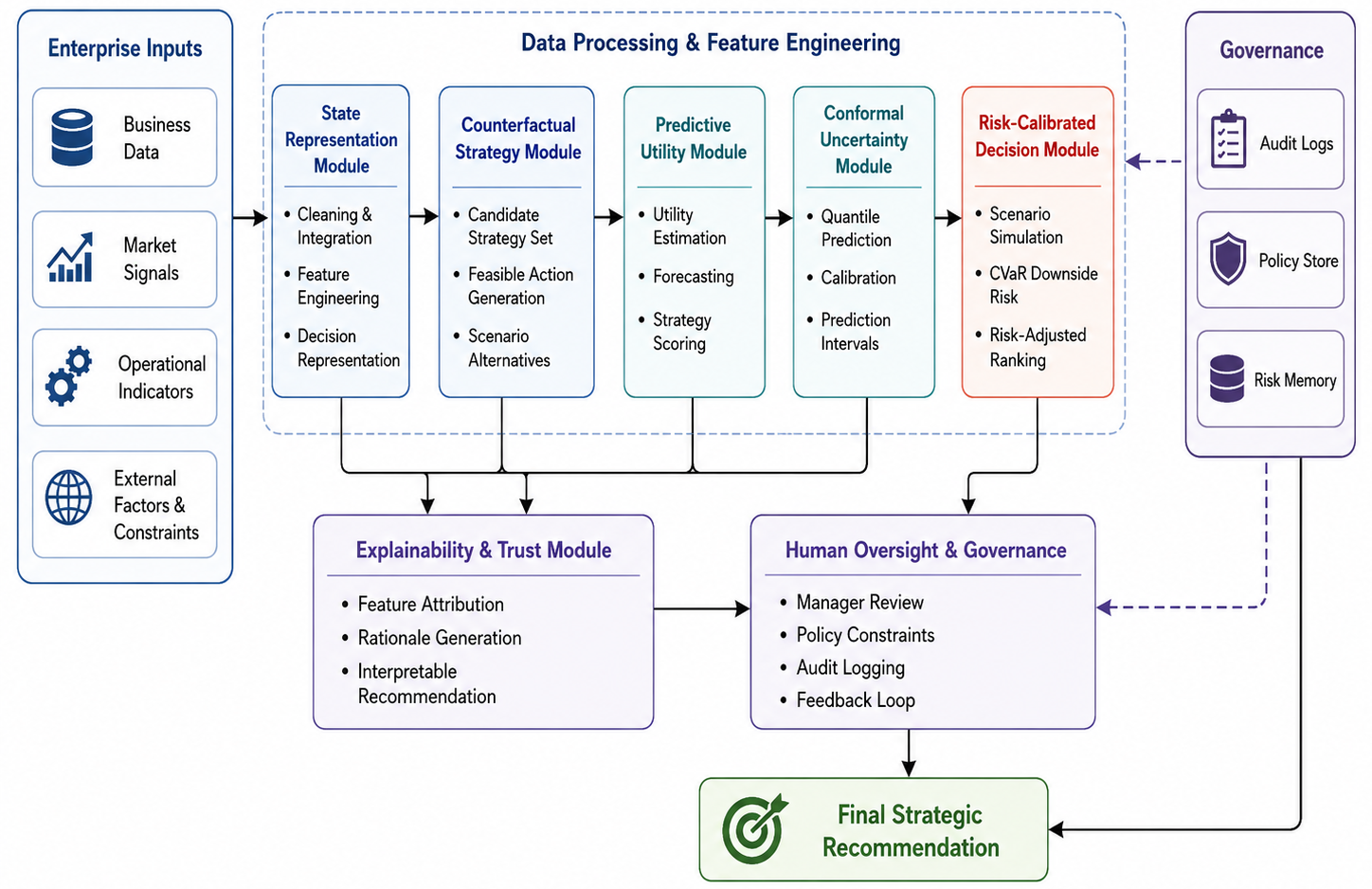}
    \caption{TRUST-ESD framework for risk-aware enterprise decision support.}
    \label{figure_1}
\end{figure}

Fig.~\ref{figure_1} shows the overall TRUST-ESD pipeline. Enterprise inputs are first transformed into a decision-ready state representation, after which feasible counterfactual strategies are generated and evaluated through predictive utility estimation, conformal uncertainty calibration, CVaR-based risk scoring \cite{moghimi2025beyond}, explainability \cite{kenny2021explaining}, governance validation \cite{european2023aiact,european2016gdpr}, and human oversight \cite{aihleg2018ethics}.

The end-to-end process is summarized as
\begin{equation}
x_t
\rightarrow
z_t
\rightarrow
A_t^{\mathrm{feas}}
\rightarrow
\hat{u}_{t,k}
\rightarrow
PI_{t,k}^{\mathrm{conf}}
\rightarrow
R_{t,k}
\rightarrow
S_{t,k}
\rightarrow
y_t^{*}.
\end{equation}
The final output is a governance-aware recommendation:
\begin{equation}
y_t^{*}=\{a_t^{*},\psi_t^{*},c_t^{*},\ell_t^{*}\},
\end{equation}
where $\psi_t^{*}$ is the explanation, $c_t^{*}$ the compliance report, and $\ell_t^{*}$ the audit record.

\subsection{Counterfactual Strategy Generation and Utility Estimation}

TRUST-ESD evaluates multiple feasible strategies rather than producing a single prediction. Each strategy is represented as
\begin{equation}
a_k=[\Delta q_k,\Delta p_k,\Delta b_k,\Delta c_k],
\end{equation}
corresponding to resource, pricing, budget, and capacity adjustments. Feasible actions are obtained by
\begin{equation}
A_t^{\mathrm{feas}}
=
\{a_k\in A_t \mid c_j(z_t,a_k)=1,\ \forall c_j\in C_{\mathrm{hard}}\}.
\end{equation}
For each feasible strategy, the predictive utility model estimates
\begin{equation}
\hat{u}_{t,k}=g_{\theta}(\xi_{t,k}).
\end{equation}
The realized utility is computed as
\begin{equation}
U_{t,k}=U(z_t,a_k,d_t),
\end{equation}
where $d_t$ is the observed outcome. For inventory-oriented decisions, this can be instantiated as
\begin{equation}
U_{t,k}
=
p_t\min(q_{t,k},d_t)
-
c_tq_{t,k}
-
h_t(q_{t,k}-d_t)^{+}
-
s_t(d_t-q_{t,k})^{+}
-
C(a_k),
\end{equation}
where $(x)^{+}=\max(x,0)$.

\subsection{Conformal Uncertainty and CVaR Risk Scoring}

To avoid overconfident recommendations, TRUST-ESD estimates calibrated uncertainty for each strategy, supporting reliable and trustworthy decision-making \cite{aihleg2018ethics,jobin2019global}. Given lower, median, and upper utility quantiles

\begin{equation}
\hat{q}_{\tau_l}(\xi_{t,k}),\quad
\hat{q}_{0.5}(\xi_{t,k}),\quad
\hat{q}_{\tau_u}(\xi_{t,k}),
\end{equation}
the predicted utility is $\hat{u}_{t,k}=\hat{q}_{0.5}(\xi_{t,k})$. The conformal prediction interval is
\begin{equation}
PI_{t,k}^{\mathrm{conf}}
=
[
\hat{q}_{\tau_l}(\xi_{t,k})-\Delta_{\alpha},
\hat{q}_{\tau_u}(\xi_{t,k})+\Delta_{\alpha}
],
\end{equation}
and the uncertainty score is
\begin{equation}
\hat{\sigma}_{t,k}
=
\operatorname{Upper}(PI_{t,k}^{\mathrm{conf}})
-
\operatorname{Lower}(PI_{t,k}^{\mathrm{conf}}).
\end{equation}

For downside-risk estimation, utility scenarios are sampled from the calibrated interval:
\begin{equation}
\tilde{u}_{t,k}^{(s)}\sim PI_{t,k}^{\mathrm{conf}},
\qquad
L_{t,k}^{(s)}=-\tilde{u}_{t,k}^{(s)}.
\end{equation}
The tail risk is measured by
\begin{equation}
\operatorname{CVaR}_{\tau}(a_k)
=
\mathbb{E}
[
L_{t,k}^{(s)}
\mid
L_{t,k}^{(s)}
\geq
\operatorname{VaR}_{\tau}(a_k)
].
\end{equation}
The model-based risk score is
\begin{equation}
R_{t,k}^{\mathrm{model}}
=
\lambda_1\operatorname{CVaR}_{\tau}(a_k)
+
\lambda_2\Pr(L_{t,k}>0)
+
\lambda_3\Pr(Viol_{t,k}=1).
\end{equation}

\subsection{Risk Memory and Policy-as-Code Governance}

TRUST-ESD stores previous decisions in a risk memory
\begin{equation}
\mathcal{M}
=
\{(z_i,a_i,\hat{u}_i,u_i,R_i,V_i,H_i)\}_{i=1}^{N},
\end{equation}
where $u_i$, $R_i$, $V_i$, and $H_i$ denote realized utility, realized risk, policy violation, and human override status. For a new strategy, similar historical cases are retrieved as
\begin{equation}
\mathcal{N}_{t,k}
=
\operatorname{TopM}
(
\operatorname{sim}((z_t,a_k),(z_i,a_i)),\mathcal{M}
).
\end{equation}
The memory-based risk is
\begin{equation}
R_{t,k}^{\mathrm{mem}}
=
\frac{1}{|\mathcal{N}_{t,k}|}
\sum_{i\in\mathcal{N}_{t,k}}
[
\eta_1(\hat{u}_i-u_i)^{+}
+
\eta_2R_i
+
\eta_3V_i
+
\eta_4H_i
].
\end{equation}
The final risk estimate is
\begin{equation}
R_{t,k}
=
\delta R_{t,k}^{\mathrm{model}}
+
(1-\delta)R_{t,k}^{\mathrm{mem}}.
\end{equation}

Governance is implemented through policy-as-code rules $P=\{p_1,\ldots,p_J\}$ to support regulatory and organizational compliance \cite{european2023aiact}, where
\begin{equation}
p_j(z_t,a_k,\hat{u}_{t,k},\hat{\sigma}_{t,k},R_{t,k})\in\{0,1\}.
\end{equation}
The governance compliance score and hard-violation indicator are
\begin{equation}
G_{t,k}
=
\frac{1}{J}
\sum_{j=1}^{J}
p_j(z_t,a_k,\hat{u}_{t,k},\hat{\sigma}_{t,k},R_{t,k}),
\end{equation}
\begin{equation}
V_{t,k}
=
\mathbb{I}
[
\exists p_j\in P_{\mathrm{hard}}:
p_j(z_t,a_k,\hat{u}_{t,k},\hat{\sigma}_{t,k},R_{t,k})=0
].
\end{equation}

\subsection{Explainability, Human Oversight, and Audit Logging}

TRUST-ESD explains the selected strategy at the decision-score level \cite{kenny2021explaining}:
\begin{equation}
\psi_t^{*}=e(z_t,a_t^{*},S_{t,k}).
\end{equation}
A generic score decomposition is
\begin{equation}
S_{t,k}
=
S_0
+
\sum_{r=1}^{d}\Phi_r^{u}
-
\sum_{r=1}^{d}\Phi_r^{\sigma}
-
\sum_{r=1}^{d}\Phi_r^{R}
+
\Phi^{G},
\end{equation}
where the terms represent utility, uncertainty, risk, and governance contributions. The final recommendation is validated through
\begin{equation}
y_t^{*}
=
h(a_t^{*},\psi_t^{*},G_{t,k},R_{t,k},M_t),
\end{equation}
where $M_t\in\{\mathrm{accept},\mathrm{revise},\mathrm{override}\}$ denotes managerial review. Each decision creates an audit record
\begin{equation}
\ell_t
=
\{x_t,A_t,a_t^{*},S_{t,k},\hat{u}_{t,k},\hat{\sigma}_{t,k},R_{t,k},G_{t,k},\psi_t^{*},M_t\},
\end{equation}
which is stored for accountability and used to update the risk memory:
\begin{equation}
\mathcal{M}\leftarrow \mathcal{M}\cup\{\ell_t\}.
\end{equation}

Algorithm~\ref{alg:trust-esd} summarizes the end-to-end TRUST-ESD pipeline from enterprise state representation to governance-aware recommendation.

\begin{algorithm}[htbp]
\caption{TRUST-ESD Risk-Calibrated Governance-Aware Decision}
\label{alg:trust-esd}
\begin{algorithmic}[1]
\Require Enterprise state $x_t$, candidate strategies $A_t$, policy set $P$, risk memory $\mathcal{M}$
\Ensure Final recommendation $y_t^{*}$

\State Construct enterprise representation $z_t=\phi(x_t)$
\State Generate feasible strategies $A_t^{\mathrm{feas}}$ using operational constraints

\For{each strategy $a_k\in A_t^{\mathrm{feas}}$}
    \State Construct state--action representation $\xi_{t,k}=\Gamma(z_t,a_k)$
    \State Estimate predicted utility $\hat{u}_{t,k}=g_{\theta}(\xi_{t,k})$
    \State Compute conformal interval $PI_{t,k}^{\mathrm{conf}}$ and uncertainty $\hat{\sigma}_{t,k}$
    \State Simulate utility scenarios and estimate $R_{t,k}^{\mathrm{model}}$ using CVaR
    \State Retrieve similar historical cases from $\mathcal{M}$
    \State Compute memory-based risk $R_{t,k}^{\mathrm{mem}}$
    \State Combine risks $R_{t,k}=\delta R_{t,k}^{\mathrm{model}}+(1-\delta)R_{t,k}^{\mathrm{mem}}$
    \State Evaluate governance compliance $G_{t,k}$ and violation indicator $V_{t,k}$
    \State Compute TRUST-ESD score $S_{t,k}$
    \State Generate explanation $\psi_{t,k}$
\EndFor

\State Select $a_t^{*}=\arg\max_{a_k\in A_t^{\mathrm{feas}}}S_{t,k}$ subject to hard policy satisfaction
\State Validate recommendation through human review if required
\State Generate final recommendation $y_t^{*}=\{a_t^{*},\psi_t^{*},c_t^{*},\ell_t^{*}\}$
\State Update audit log and risk memory $\mathcal{M}$
\State \Return $y_t^{*}$
\end{algorithmic}
\end{algorithm}

\section{Experimental Process}

\subsection{Datasets and Training Processes}

TRUST-ESD is evaluated in an enterprise decision-support setting where the goal is to select risk-calibrated and governance-compliant strategies, following the full pipeline from state representation and strategy generation to uncertainty calibration, CVaR risk scoring \cite{moghimi2025beyond}, policy validation \cite{european2023aiact,european2016gdpr}, explanation \cite{kenny2021explaining}, and human oversight \cite{aihleg2018ethics}.

Given an enterprise state $x_t$ and a feasible strategy set
\begin{equation}
A_t^{\mathrm{feas}}=\{a_1,a_2,\ldots,a_K\},
\end{equation}
each method produces a recommendation by ranking candidate strategies. Prediction-only baselines select
\begin{equation}
a_t^{\mathrm{pred}}
=
\arg\max_{a_k\in A_t^{\mathrm{feas}}}
\hat{u}_{t,k},
\end{equation}
whereas TRUST-ESD selects
\begin{equation}
a_t^{*}
=
\arg\max_{a_k\in A_t^{\mathrm{feas}}}
S_{t,k},
\end{equation}
where $S_{t,k}$ jointly accounts for predicted utility, calibrated uncertainty, downside risk \cite{moghimi2025beyond}, risk memory, and governance compliance \cite{jobin2019global,european2023aiact}.

The experimental benchmark is built from enterprise-oriented retail tabular data, using \textit{Rossmann Store Sales} \cite{rossmann-store-sales} and \textit{M5 Forecasting} \cite{m5-forecasting-accuracy} to simulate strategic decision support. Each instance represents a store--item--time state, where candidate strategies include conservative replenishment, aggressive replenishment, promotion, cost-saving, risk-control, and capacity expansion. Realized utility, downside risk, and policy compliance are computed using observed demand or sales. The data are chronologically split into training, validation, calibration, and test sets for model fitting, hyperparameter tuning, conformal calibration, and final evaluation, respectively, with all preprocessing fitted only on the training set to avoid information leakage.

\subsection{Baseline Models}

TRUST-ESD is evaluated against prediction-only, uncertainty-aware, and decision-aware baselines. The prediction-only group includes Linear Regression \cite{tripepi2008linear}, Random Forest \cite{salman2024random}, XGBoost \cite{miraj2025gksmote}, LightGBM \cite{tyralis2023merging}, CatBoost \cite{miraj2025gksmote}, TabNet, and FT-Transformer \cite{mota2025comparative}, which select strategies by maximizing predicted utility. Comparative machine learning studies further support the use of integrated and nonlinear models for complex prediction tasks \cite{zhao2025enhanced}. The uncertainty-aware group includes NGBoost \cite{miraj2025gksmote}, Quantile LightGBM \cite{tyralis2023merging}, and Deep Ensembles \cite{abe2022deep}, which additionally estimate predictive uncertainty but do not directly optimize downside risk \cite{moghimi2025beyond} or governance compliance \cite{european2023aiact}. The decision-aware group includes Utility-Only, Uncertainty-Penalized, CVaR-Risk \cite{moghimi2025beyond}, Policy-Filtered \cite{european2023aiact,european2016gdpr}, TRUST-ESD variants enable controlled comparison of utility, uncertainty, risk, policy filtering, memory, and the full framework.

\subsection{Evaluation Metrics}

The evaluation covers prediction quality, decision quality, risk control, calibration, explainability, and governance. Predictive performance is measured using RMSE, MAE, and $R^2$ for regression tasks \cite{yang2026deadmap}, and Accuracy \cite{miraj2025gksmote,zhao2025covert}, F1-score \cite{miraj2025gksmote,yang2025entity}, and AUC \cite{yang2026deadmap,yang2025entity}for classification tasks.

Decision quality is evaluated using Average Utility and Risk-Adjusted Utility:
\begin{equation}
AU
=
\frac{1}{N}
\sum_{t=1}^{N}
U_{t,a_t},
\end{equation}
\begin{equation}
RAU
=
\frac{1}{N}
\sum_{t=1}^{N}
\left(
U_{t,a_t}
-
\rho R_{t,a_t}
\right),
\end{equation}
where $U_{t,a_t}$ is the realized utility of the selected strategy and $R_{t,a_t}$ is its risk score.

Risk-control performance is measured using Average Risk Exposure, Downside Loss, and CVaR \cite{moghimi2025beyond}:
\begin{equation}
DL
=
\frac{1}{N}
\sum_{t=1}^{N}
\max(0,-U_{t,a_t}).
\end{equation}

Uncertainty reliability is evaluated using Prediction Interval Coverage Probability and Mean Prediction Interval Width, which assess whether calibrated prediction intervals provide reliable uncertainty estimates:
\begin{equation}
PICP
=
\frac{1}{N}
\sum_{t=1}^{N}
\mathbb{I}
[
u_t \in PI_t^{\mathrm{conf}}
],
\end{equation}
\begin{equation}
MPIW
=
\frac{1}{N}
\sum_{t=1}^{N}
\left(
Upper(PI_t^{\mathrm{conf}})
-
Lower(PI_t^{\mathrm{conf}})
\right).
\end{equation}

Governance performance is measured using Governance Compliance Rate, Policy Violation Rate, Human Review Rate, Human Override Rate, and Audit Completeness, reflecting trustworthy AI requirements for oversight \cite{aihleg2018ethics}, accountability \cite{barletta2023responsible,jobin2019global}, regulatory compliance \cite{european2023aiact,european2016gdpr}, and secure machine learning deployment \cite{enisa2021securing}:
\begin{equation}
GCR
=
\frac{\#\text{compliant recommendations}}
{\#\text{total recommendations}},
\qquad
PVR=1-GCR.
\end{equation}

Explainability is measured by Fidelity and Attribution Stability to evaluate whether TRUST-ESD provides reliable and interpretable strategic recommendations \cite{kenny2021explaining}.

\subsection{Candidate Strategy Construction and Implementation Details}

To convert predictive outputs into decision-support instances, each test sample is paired with the same finite candidate strategy set as in Table~\ref{tab:candidate_strategies}. In the retail decision setting, based on demand-forecasting benchmarks such as \textit{Rossmann Store Sales} \cite{rossmann-store-sales} and \textit{M5 Forecasting} \cite{m5-forecasting-accuracy}, each strategy defines an operational quantity $q_{t,k}$ using the predicted demand $\hat{d}_t$ and the calibrated prediction interval
$PI_t^{\mathrm{conf}}=[l_t,u_t]$. This ensures fair evaluation under identical actions, outcomes, and constraints.

\begin{table}[htbp]
\centering
\caption{Candidate strategy definitions used for decision evaluation.}
\label{tab:candidate_strategies}
\begin{tabular}{lll}
\hline
Strategy & Action Rule & Main Risk Source \\
\hline
Conservative & $q_{t,k}=1.05\hat{d}_t$ & mild overstock \\
Aggressive & $q_{t,k}=1.25\hat{d}_t$ & high holding cost \\
Cost-saving & $q_{t,k}=0.90\hat{d}_t$ & stockout risk \\
Promotion & $q_{t,k}=1.15\hat{d}_t$ & promotion cost \\
Risk-control & $q_{t,k}=u_t$ & higher inventory cost \\
Capacity expansion & $q_{t,k}=\min(1.30\hat{d}_t,q_{\max})$ & fixed expansion cost \\
\hline
\end{tabular}
\end{table}

For each selected strategy, realized utility is computed using the observed outcome $d_t$, enabling evaluation beyond prediction error through decision utility, downside loss, CVaR-based risk control \cite{moghimi2025beyond}, calibration quality, and policy compliance \cite{european2023aiact,european2016gdpr}. All models use the same preprocessing, training, calibration, and test protocol to ensure a fair comparison: hyperparameters are tuned on the validation set, $\Delta_{\alpha}$ is estimated on the calibration set, and the test set is used only for final reporting. The TRUST-ESD coefficients $\alpha,\beta,\gamma,\omega,\mu$, scenario number $S$, CVaR level $\tau$ \cite{moghimi2025beyond}, and memory size $M$ are selected by validation performance under hard policy constraints. Governance rules are implemented as deterministic constraints aligned with trustworthy AI oversight \cite{aihleg2018ethics}, accountability-oriented AI governance \cite{jobin2019global,barletta2023responsible}, and secure machine learning deployment principles \cite{enisa2021securing}, while results are reported as mean values with standard deviations over repeated runs.

\section{Results and Analysis}
\subsection{Overall Performance}
Full TRUST-ESD achieves the strongest overall decision-support performance while maintaining competitive predictive accuracy. As shown in Table~\ref{tab:overall_performance}, Deep Ensembles obtains the best RMSE and MAE, but its decision quality and trustworthiness metrics are weaker than those of the proposed framework. Full TRUST-ESD reaches the highest RAU of $0.815$, lowest risk exposure of $0.162$, lowest CVaR of $0.218$, lowest ECE of $0.031$, highest PICP of $0.939$, highest Fidelity of $0.834$, and highest GCR of $0.967$. Compared with Deep Ensembles, TRUST-ESD improves RAU by $7.95\%$, reduces risk exposure by $23.22\%$, reduces CVaR by $23.78\%$, lowers ECE by $13.89\%$, improves PICP by $1.19\%$, improves explanation fidelity by $10.90\%$, and improves governance compliance by $9.76\%$. 

\begin{table*}[htbp]
\centering
\caption{Overall performance results. Best in \textbf{bold}, and second-best in \underline{underlined}.} 
\label{tab:overall_performance}
\resizebox{\textwidth}{!}{
\begin{tabular}{lcccccccccc}
\hline
Model 
& RMSE$\downarrow$ 
& MAE$\downarrow$ 
& Utility$\uparrow$ 
& RAU$\uparrow$ 
& Risk Exp.$\downarrow$ 
& CVaR$\downarrow$ 
& ECE$\downarrow$ 
& PICP$\uparrow$ 
& Fidelity$\uparrow$ 
& GCR$\uparrow$ \\
\hline
Linear Regression 
& 0.168 & 0.124 & 0.701 & 0.612 & 0.318 & 0.421 & 0.091 & 0.812 & 0.642 & 0.804 \\

Random Forest 
& 0.143 & 0.105 & 0.742 & 0.651 & 0.292 & 0.384 & 0.076 & 0.845 & 0.681 & 0.823 \\

XGBoost 
& 0.129 & 0.094 & 0.781 & 0.692 & 0.263 & 0.352 & 0.063 & 0.872 & 0.713 & 0.841 \\

LightGBM 
& 0.124 & 0.091 & 0.804 & 0.711 & 0.248 & 0.336 & 0.058 & 0.884 & 0.729 & 0.852 \\

CatBoost 
& 0.122 & 0.089 & 0.811 & 0.718 & 0.241 & 0.327 & 0.056 & 0.887 & 0.736 & 0.858 \\

TabNet 
& 0.135 & 0.098 & 0.776 & 0.684 & 0.257 & 0.348 & 0.066 & 0.864 & 0.754 & 0.846 \\

FT-Transformer 
& \underline{0.121} & \underline{0.088} & 0.817 & 0.731 & 0.232 & 0.314 & 0.052 & 0.891 & 0.768 & 0.867 \\

NGBoost 
& 0.131 & 0.095 & 0.795 & 0.734 & 0.226 & 0.301 & 0.041 & 0.915 & 0.724 & 0.861 \\

Quantile LightGBM 
& 0.126 & 0.092 & 0.806 & 0.746 & 0.219 & 0.294 & 0.038 & 0.923 & 0.731 & 0.873 \\

Deep Ensembles 
& \textbf{0.119} & \textbf{0.086} & \underline{0.823} & 0.755 & 0.211 & 0.286 & 0.036 & 0.928 & 0.752 & 0.881 \\

Utility-Only 
& 0.122 & 0.089 & \textbf{0.836} & 0.704 & 0.284 & 0.369 & 0.057 & 0.866 & 0.741 & 0.812 \\

Uncertainty-Penalized 
& 0.122 & 0.089 & 0.821 & 0.758 & 0.224 & 0.302 & 0.039 & 0.921 & 0.749 & 0.846 \\

CVaR-Risk 
& 0.123 & 0.090 & 0.812 & 0.781 & 0.196 & 0.263 & 0.038 & 0.919 & 0.755 & 0.867 \\

Policy-Filtered 
& 0.123 & 0.090 & 0.804 & 0.776 & 0.188 & 0.257 & 0.039 & 0.918 & 0.762 & 0.932 \\

TRUST-ESD w/o Risk Memory 
& 0.123 & 0.090 & 0.814 & \underline{0.793} & \underline{0.179} & \underline{0.241} & \underline{0.035} & \underline{0.931} & \underline{0.811} & \underline{0.944} \\

\textbf{Full TRUST-ESD}
& 0.124 & 0.091 & 0.818 & \textbf{0.815} & \textbf{0.162} & \textbf{0.218} & \textbf{0.031} & \textbf{0.939} & \textbf{0.834} & \textbf{0.967} \\
\hline
\end{tabular}
}
\end{table*}

\subsection{Decision Utility and Risk-Adjusted Utility}
The Utility-Only baseline achieves the highest raw utility of $0.836$, whereas Full TRUST-ESD obtains a slightly lower raw utility of $0.818$, corresponding to only a $2.15\%$ reduction. However, after risk adjustment, TRUST-ESD improves RAU from $0.704$ to $0.815$ over Utility-Only, yielding a $15.77\%$ gain in risk-adjusted decision quality. This trade-off is visualized in Fig.~\ref{Figure_2}, where its preserves high utility while substantially improving risk-adjusted value. Compared with CVaR-Risk and Policy-Filtered baselines, TRUST-ESD further improves RAU by $4.35\%$ and $5.03\%$, respectively, demonstrating that the full framework provides a stronger balance between expected value, uncertainty, risk, and governance.

\subsection{Risk Exposure and CVaR Analysis}
\begin{figure*}[htbp]
    \centering
    
    \begin{minipage}{0.48\textwidth}
        \centering
        \includegraphics[width=\linewidth]{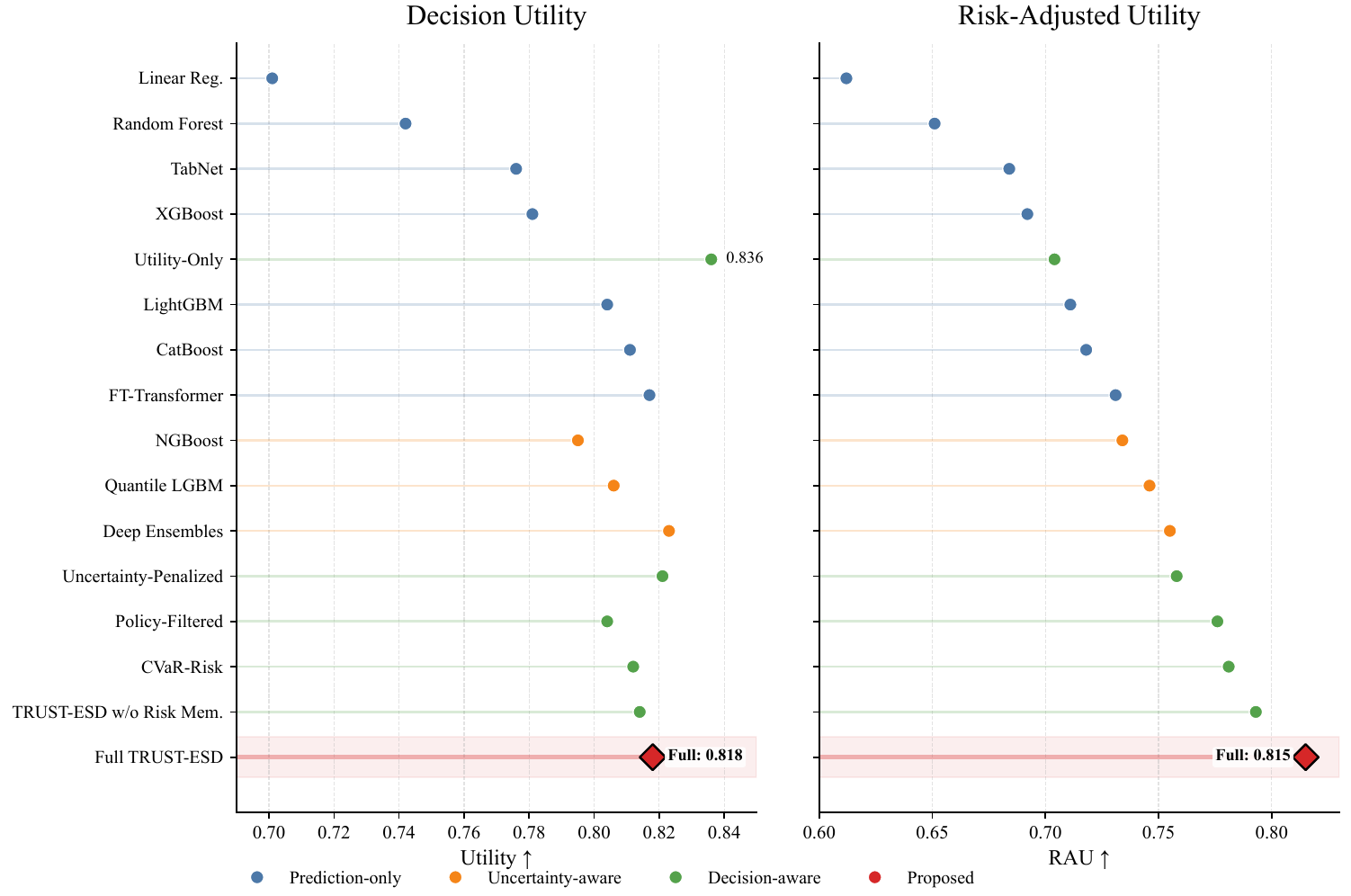}
        \caption{Decision and Risk-adjusted utility}
        \label{Figure_2}
    \end{minipage}
    \hfill
    \begin{minipage}{0.48\textwidth}
        \centering
        \includegraphics[width=\linewidth]{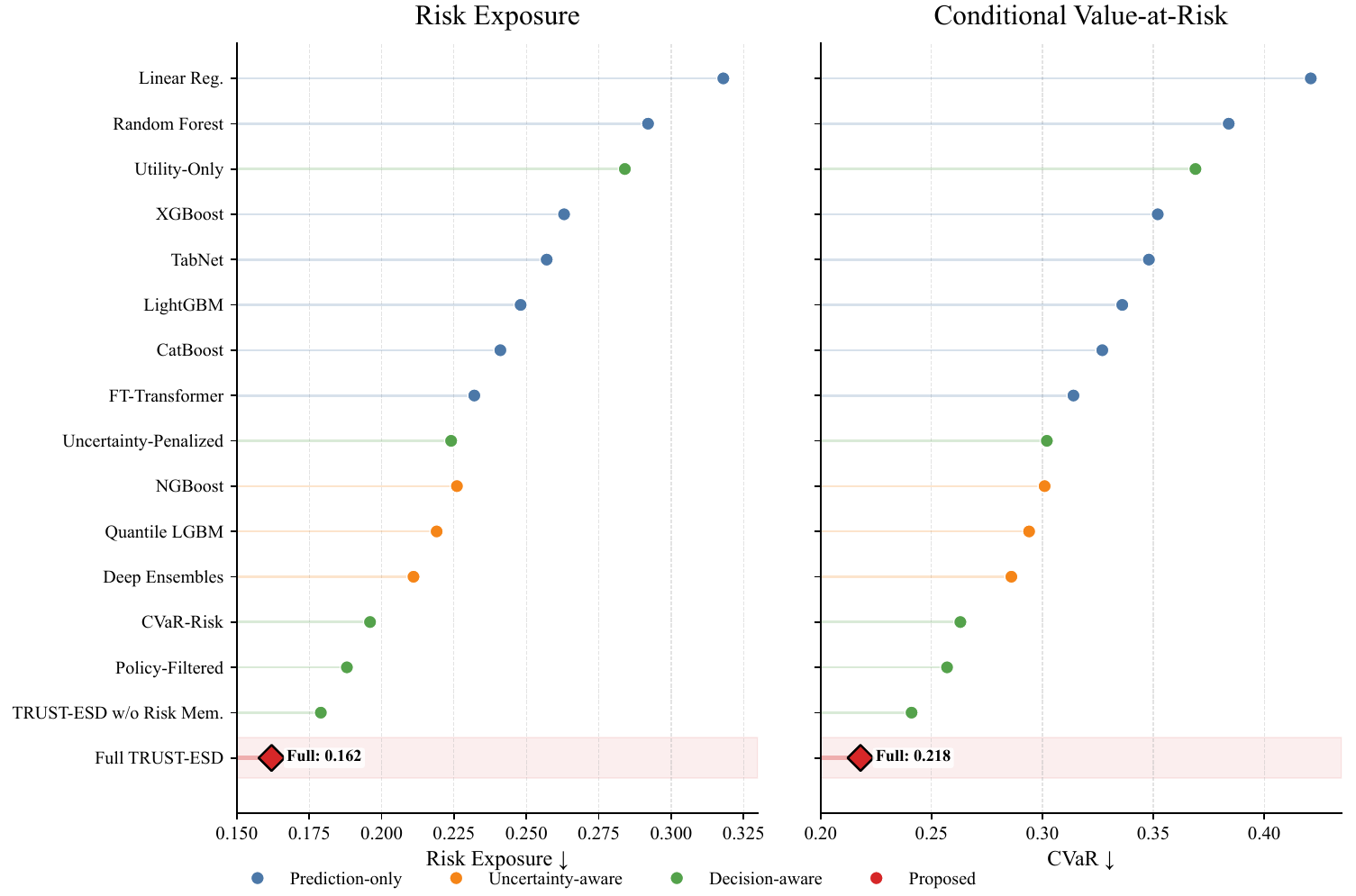}
        \caption{Risk exposure and CVaR trade-off}
        \label{Figure_3}
    \end{minipage}
\end{figure*}

TRUST-ESD provides the strongest downside-risk control. It achieves the lowest risk exposure of $0.162$ and the lowest CVaR of $0.218$, indicating that its selected strategies are less vulnerable to adverse outcomes. Compared with Policy-Filtered, TRUST-ESD reduces risk exposure by $13.83\%$ and CVaR by $15.18\%$; compared with CVaR-Risk, it reduces risk exposure by $17.35\%$ and CVaR by $17.11\%$. The advantage becomes more substantial against Utility-Only, where TRUST-ESD reduces risk exposure by $42.96\%$ and CVaR by $40.92\%$.Fig.~\ref{Figure_3} shows that TRUST-ESD achieves the best risk--utility trade-off.

\subsection{Calibration Analysis}
\begin{figure*}[htbp]
    \centering

    \begin{minipage}{0.48\textwidth}
        \centering
        \includegraphics[width=\linewidth]{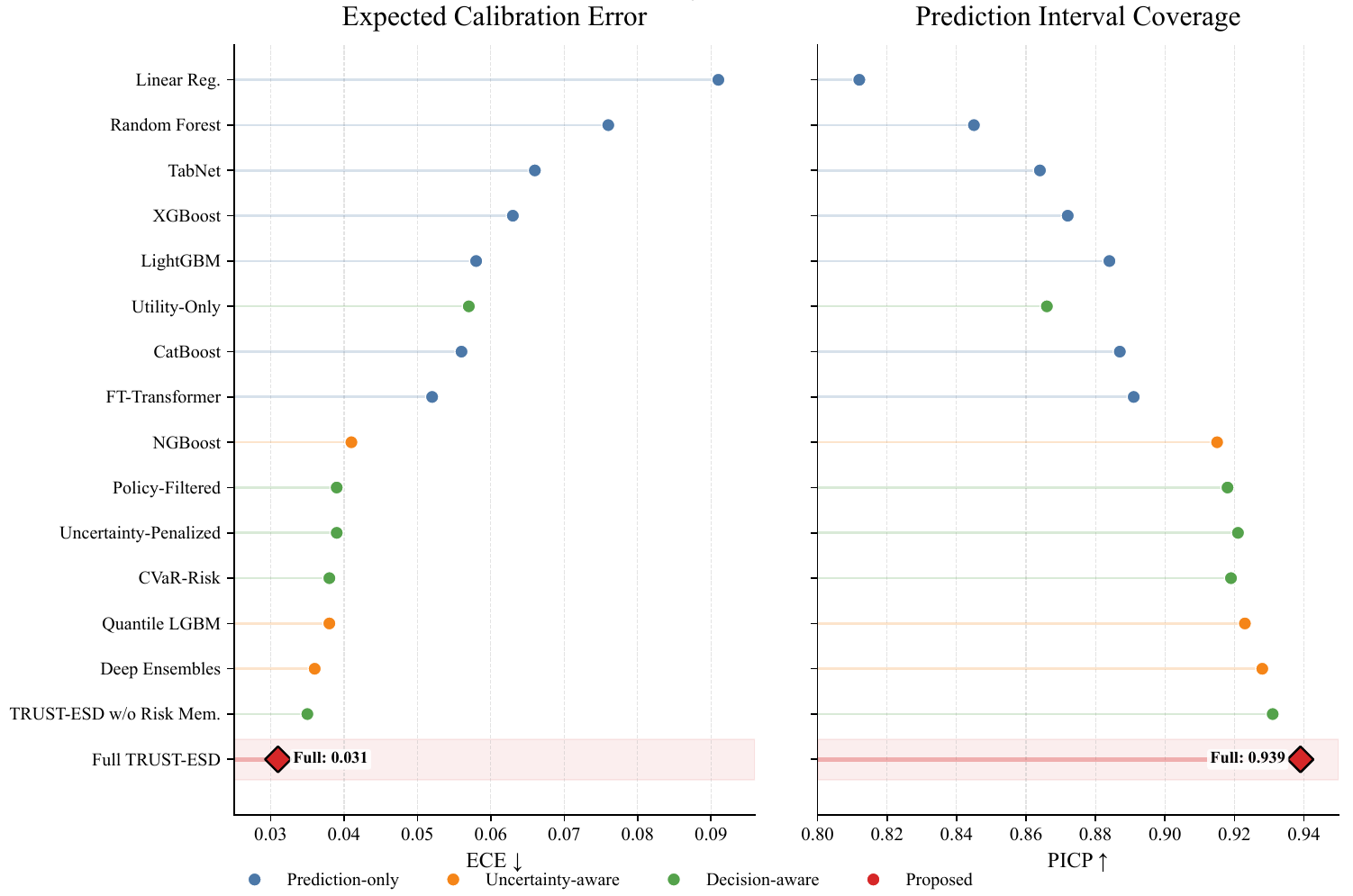}
        \caption{Calibration performance}
        \label{Figure_4}
    \end{minipage}
    \hfill
    \begin{minipage}{0.48\textwidth}
        \centering
        \includegraphics[width=\linewidth]{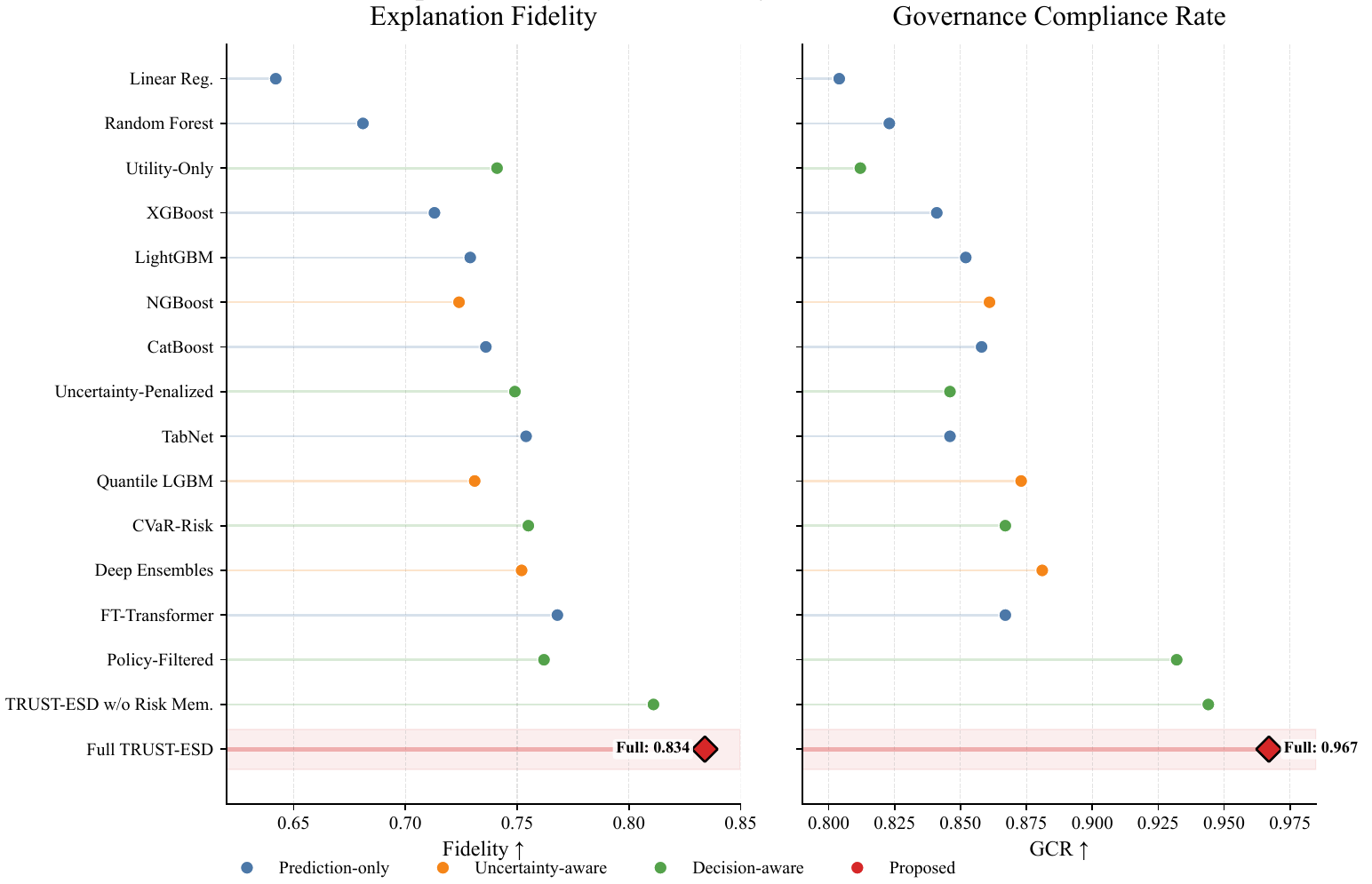}
        \caption{Explainability and Trust analysis}
        \label{Figure_5}
    \end{minipage}
\end{figure*}

Reliable uncertainty estimation is essential because overconfident recommendations can lead to costly enterprise decisions. TRUST-ESD achieves the lowest ECE of $0.031$ and the highest PICP of $0.939$, indicating both stronger calibration and better interval coverage. The calibration comparison in Fig.~\ref{Figure_4} shows that conformal uncertainty calibration improves reliability over strong uncertainty-aware baselines. Compared with Quantile LightGBM, TRUST-ESD reduces ECE by $18.42\%$ and improves PICP by $1.73\%$; compared with Deep Ensembles, it reduces ECE by $13.89\%$ and improves PICP by $1.19\%$. 

\subsection{Explainability and Trust Analysis}
Full TRUST-ESD obtains the highest explanation fidelity of $0.834$, outperforming FT-Transformer by $8.59\%$, Policy-Filtered by $9.45\%$, and Deep Ensembles by $10.90\%$. This improvement is important because TRUST-ESD explains the final strategic score, not only the predictive utility estimate. As reported in Fig.~\ref{Figure_5}, the proposed method provides stronger explainability and trust performance than both prediction-only and uncertainty-aware baselines. The ablation results further support this finding: removing the explainability module reduces fidelity from $0.834$ to $0.684$, corresponding to a $17.99\%$ degradation.

\subsection{Governance Compliance Analysis}
Governance validation substantially improves deployment suitability under enterprise policy constraints. Full TRUST-ESD achieves the highest GCR of $0.967$, improving over Policy-Filtered by $3.76\%$, Deep Ensembles by $9.76\%$, and Utility-Only by $19.09\%$. In terms of policy violation rate, TRUST-ESD reduces violations from $0.068$ to $0.033$ compared with Policy-Filtered, a $51.47\%$ reduction, and from $0.188$ to $0.033$ compared with Utility-Only, an $82.45\%$ reduction. These results, summarized visually in Fig.~\ref{Figure_6}, show that policy-as-code validation and human-governance routing help convert model outputs into safer and more compliant enterprise recommendations.

\subsection{Ablation Study}
The ablation results confirm that each TRUST-ESD component contributes to a distinct dimension of trustworthy decision support. Removing CVaR risk scoring causes a $9.08\%$ RAU degradation and increases risk exposure by $56.79\%$, showing that downside-risk modeling is central to robust strategy selection. Removing conformal calibration causes a $5.77\%$ RAU drop, increases ECE by $96.77\%$, and reduces PICP by $8.20\%$, confirming the importance of calibrated uncertainty. Risk memory also contributes meaningfully: removing it reduces RAU by $2.70\%$ and increases CVaR by $10.55\%$. As shown in Table~\ref{tab:ablation}, removing policy governance mainly reduces GCR by $9.82\%$, while removing explainability sharply lowers fidelity. The Utility-Only variant achieves the highest raw utility but suffers the largest RAU degradation of $13.62\%$, with $75.31\%$ higher risk exposure and $69.27\%$ higher CVaR than Full TRUST-ESD.

\begin{table*}[htbp]
\centering
\caption{Ablation analysis. Best in \textbf{bold}, and second-best in \underline{underlined}.}
\label{tab:ablation}
\resizebox{\textwidth}{!}{
\begin{tabular}{lccccccccc}
\hline
Variant 
& Utility$\uparrow$ 
& RAU$\uparrow$ 
& $\Delta$RAU$\downarrow$ 
& Risk Exp.$\downarrow$ 
& CVaR$\downarrow$ 
& ECE$\downarrow$ 
& PICP$\uparrow$ 
& Fidelity$\uparrow$ 
& GCR$\uparrow$ \\
\hline
TRUST-ESD w/o Conformal Calibration 
& 0.812 & 0.768 & 5.77\% & 0.191 & 0.259 & 0.061 & 0.862 & 0.819 & 0.948 \\

TRUST-ESD w/o CVaR Risk 
& 0.826 & 0.741 & 9.08\% & 0.254 & 0.338 & 0.036 & 0.928 & 0.821 & 0.951 \\

TRUST-ESD w/o Risk Memory 
& 0.814 & \underline{0.793} & \underline{2.70\%} & \underline{0.179} & \underline{0.241} & \underline{0.035} & \underline{0.931} & \underline{0.811} & \underline{0.944} \\

TRUST-ESD w/o Policy Governance 
& 0.822 & 0.776 & 4.79\% & 0.185 & 0.249 & 0.034 & 0.930 & 0.823 & 0.872 \\

TRUST-ESD w/o Explainability 
& 0.820 & 0.786 & 3.56\% & 0.174 & 0.232 & 0.033 & 0.936 & 0.684 & 0.938 \\

TRUST-ESD Utility-Only Scoring 
& \textbf{0.836} & 0.704 & 13.62\% & 0.284 & 0.369 & 0.057 & 0.866 & 0.741 & 0.812 \\

Full TRUST-ESD 
& \underline{0.818} & \textbf{0.815} & \textbf{0.00\%} & \textbf{0.162} & \textbf{0.218} & \textbf{0.031} & \textbf{0.939} & \textbf{0.834} & \textbf{0.967} \\
\hline
\end{tabular}
}
\end{table*}

\subsection{Case Study}
\begin{figure*}[htbp]
    \centering

    \begin{minipage}{0.48\textwidth}
        \centering
        \includegraphics[width=\linewidth]{fig_6_5_explainability_trust_cropped.pdf}
        \caption{Governance compliance analysis}
        \label{Figure_6}
    \end{minipage}
    \hfill
    \begin{minipage}{0.50\textwidth}
        \centering
        \includegraphics[width=\linewidth]{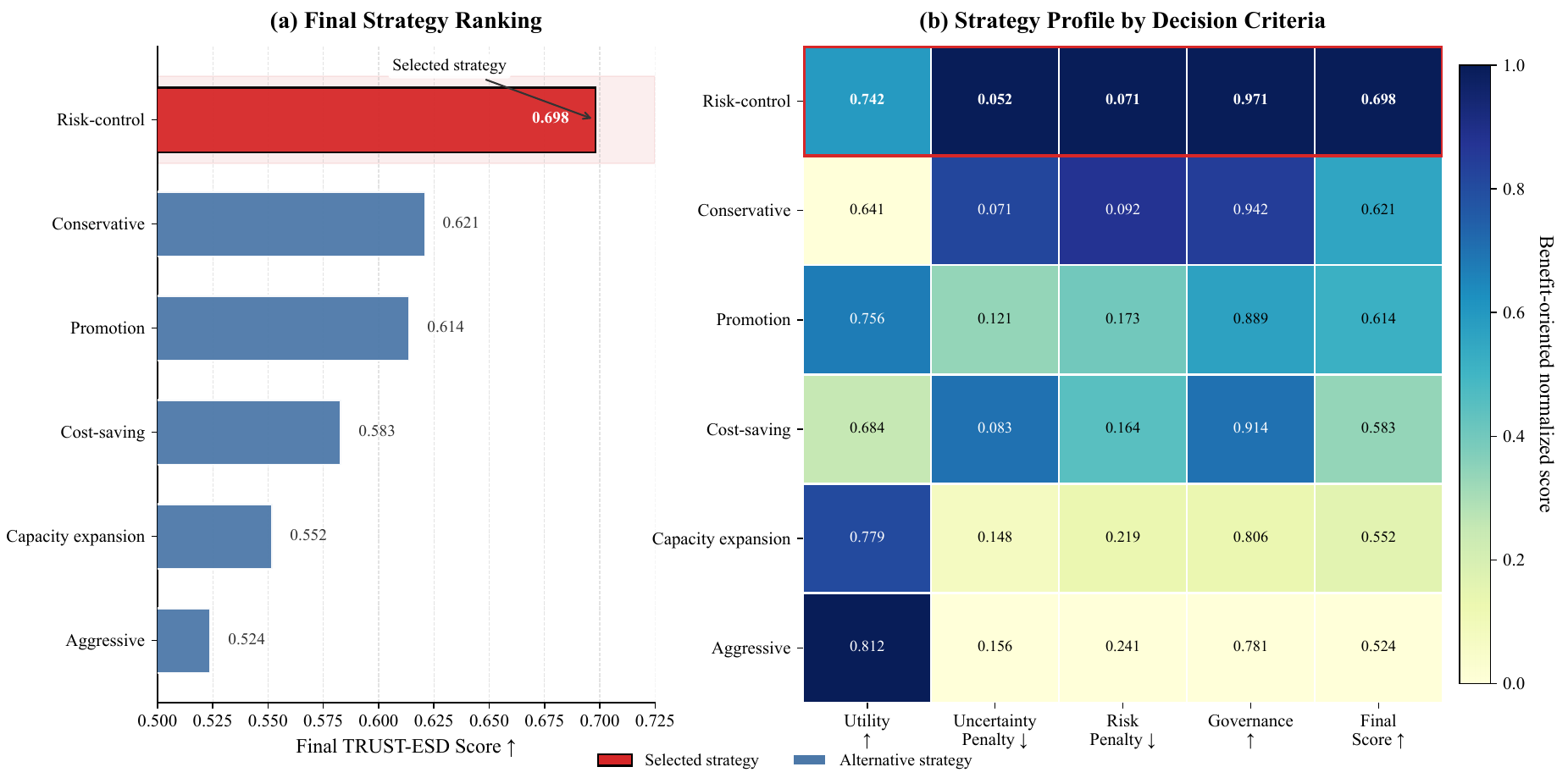}
        \caption{Case-level strategy analysis}
        \label{Figure_7}
    \end{minipage}
\end{figure*}

The case-level analysis shows how TRUST-ESD avoids aggressive but fragile strategies. The Aggressive strategy achieves the highest raw utility of $0.812$, but its high uncertainty penalty of $0.156$, high risk penalty of $0.241$, and lower governance score of $0.781$ reduce its final score to $0.524$. In contrast, the Risk-Control strategy has a slightly lower utility of $0.742$, but it reduces the uncertainty penalty by $66.67\%$, reduces the risk penalty by $70.54\%$, improves governance compliance by $24.33\%$, and increases the final decision score by $33.21\%$ compared with the Aggressive strategy. Fig.~\ref{Figure_7} shows that TRUST-ESD selects the most balanced strategy, not merely the highest-utility one.

\section{Conclusion}
This paper presented TRUST-ESD, a risk-calibrated and governance-aware framework for enterprise strategic decision support under uncertainty. By integrating predictive utility estimation, conformal uncertainty calibration, CVaR-based downside-risk scoring, risk-memory retrieval, policy-as-code governance, explainability, and human oversight, TRUST-ESD moves beyond prediction-only recommendation toward trustworthy strategic decision-making. Experimental results show that the proposed framework improves risk-adjusted utility, reduces risk exposure and CVaR, strengthens calibration, and improves explanation fidelity and governance compliance while maintaining competitive predictive accuracy. The ablation and case-study analyses further confirm that each component contributes to robust, interpretable, and policy-compliant decision support. Future work will extend TRUST-ESD to dynamic enterprise environments, adaptive governance policies, and real-world human feedback settings.


%
%
%
%

\end{document}